\documentclass[letterpaper, 10 pt, conference]{ieeeconf}  

\IEEEoverridecommandlockouts                              

\usepackage{cite}
\usepackage{hyperref}
\usepackage{amsmath,amssymb,amsfonts}
\usepackage{algorithmic}
\usepackage{graphicx}
\usepackage{textcomp}
\usepackage{xcolor}
\usepackage{amsmath}
\usepackage{booktabs}
\usepackage{multirow}
\usepackage{graphicx} 
\usepackage{hyperref}
\usepackage{tabularx}
\usepackage{amssymb}
\def\BibTeX{{\rm B\kern-.05em{\sc i\kern-.025em b}\kern-.08em
    T\kern-.1667em\lower.7ex\hbox{E}\kern-.125emX}}

\title{\LARGE \bf
Cross-Modal Iteration Distillation for Robust IHD Screening: The IDNet Framework and A New Benchmark
}

\author{
    Yongchang Gao$^{1,*}$, Junjie Pang$^{1,*}$, Shuaiyu Yang$^{1,*}$, \\
    Yusheng Yang$^{2}$, Xichao Jia$^{1}$, Shaojie Li$^{1}$, Hongfei Zhang$^{3}$, and Jia Mu$^{2,\dagger}$%
    \thanks{*These authors contributed equally to this work.}%
    \thanks{$^{\dagger}$Corresponding author.}%
    \thanks{$^{1}$Yongchang Gao, Junjie Pang, Shuaiyu Yang, Xichao Jia, and Shaojie Li are with the University of Chinese Academy of Sciences, Beijing, China (e-mail: \{gaoyongchang24, pangjunjie25, yangshuaiyu24, jiaxichao24\}@mails.ucas.ac.cn).}%
    \thanks{$^{2}$Yusheng Yang and Jia Mu are with MGI Tech Co., Ltd., Shenzhen, China (e-mail: \{yangyusheng, mujia\}@genomics.cn).}%
    \thanks{$^{3}$Hongfei Zhang is with Shenzhen University, Shenzhen, China (e-mail: 2510232028@mails.szu.edu.cn).}%
}

\begin{document}

\maketitle
\thispagestyle{empty}
\pagestyle{empty}

\begin{abstract}
Color Fundus Photography (CFP) offers a low-cost and non-invasive route for ischemic heart disease (IHD) screening, but current studies are limited by scarce public benchmarks and ineffective fusion of retinal images with sparse clinical variables. We propose IDNet, a multimodal framework with a Cross-Modal Distillation Aggregator (CDA) that uses learnable queries to sequentially integrate left-eye, right-eye, and clinical features, mitigating the imbalance between high-dimensional visual features and low-dimensional tabular inputs. We also construct a reproducible UK Biobank benchmark with open-source curation and quality-control pipelines, yielding 50,410 images from 25,205 subjects. On this benchmark, IDNet outperforms image-only, clinical-only, and several multimodal baselines, and CDA consistently improves multiple visual encoders as a plug-in fusion module.

\end{abstract}

\section{Introduction}
\label{sec:intro}
Ischemic Heart Disease (IHD) remains the leading global cause of mortality, necessitating accessible early screening strategies~\cite{li2024systematic}. While Coronary CT Angiography (CTCA) serves as the diagnostic gold standard, its high cost and radiation risks limit widespread community deployment. Conversely, the retina offers a unique, non-invasive window into systemic vascular health~\cite{mcgeecan2009retinal}. Recent advances in deep learning, particularly foundation models like RETFound~\cite{zhou2023foundation}, have demonstrated the potential of fundus imaging for predicting cardiovascular biomarkers. However, translating these advances into robust IHD screening tools faces significant hurdles.

An important challenge lies in multimodal fusion. Existing methods either depend on extensive clinical examinations, reducing accessibility, or combine fundus images with clinical variables using simple strategies such as concatenation. This setting is difficult because retinal image features are high-dimensional and fine-grained, whereas the available clinical variables are sparse and low-dimensional~\cite{wang2025artificial}. As a result, naive fusion may underutilize clinical priors and fail to capture complementary information across binocular views and systemic risk factors.

Another limitation is the lack of standardized resources. Compared with diabetic retinopathy research, publicly available fundus datasets specifically curated for IHD screening remain limited~\cite{abramoff2016improved}~\cite{nagpal2022review}. This hinders both model development and fair comparison across studies, as existing works often rely on different cohorts, exclusion criteria, and preprocessing pipelines.

To address these challenges, we propose \textbf{IDNet}, a multimodal framework for low-cost IHD screening from binocular fundus photographs and minimal clinical covariates. The key component is a \textbf{Cross-Modal Distillation Aggregator (CDA)}, which uses learnable queries to sequentially integrate visual and clinical information. Our main contributions are as follows:

\begin{itemize}
    \item \textbf{A reproducible benchmark protocol.} We construct a UK Biobank-based benchmark and release the corresponding curation and quality-control pipeline, resulting in 50,410 images from 25,205 subjects.
    \item \textbf{A sequential multimodal fusion module.} We introduce CDA, which performs query-based cross-modal interaction over binocular fundus features and clinical covariates.
    \item \textbf{Extensive empirical validation.} IDNet improves over image-only, clinical-only, and several multimodal baselines, and CDA consistently benefits multiple visual encoders as a plug-in fusion module.
\end{itemize}

\section{Dataset Construction}

\begin{figure}[htbp]
  \centering
  \includegraphics[width=1\linewidth]{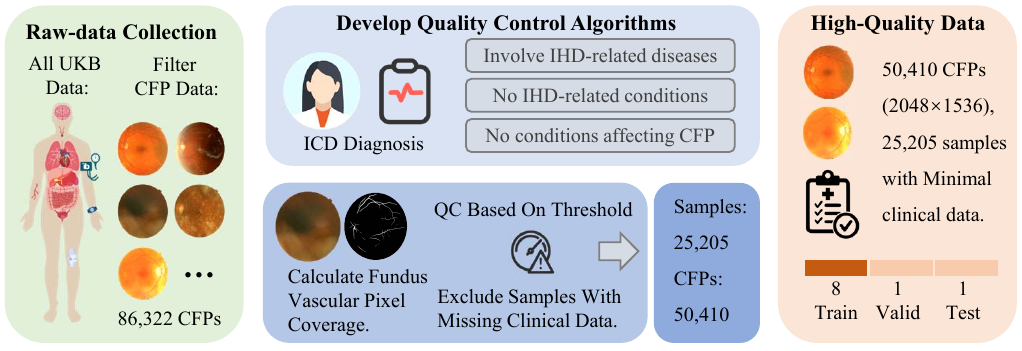} 
  \caption{The overall dataset construction pipeline. We filtered the raw UKB data through rigorous ICD-based diagnostic criteria and a threshold-based image quality control algorithm (evaluating vascular pixel coverage) to generate a high-quality benchmark dataset for IHD screening.}
  \label{fig:D_C} 
  \vspace{-1.0em}
\end{figure}

As shown in Fig.~\ref{fig:D_C}, we constructed a reproducible benchmark from the UK Biobank (UKB). Starting from 86,322 raw fundus images, we applied ICD-based diagnostic filtering, image-quality control based on vascular pixel coverage, and exclusion of samples with missing clinical covariates (age, sex, smoking, and alcohol history). The final dataset contains 50,410 images from 25,205 subjects, including 3,775 IHD-positive cases, and is split into training, validation, and test sets at an 8:1:1 ratio. The study was approved by the Institutional Review Board with waived informed consent due to its retrospective design. All dataset splits were performed at the subject level to ensure that images from the same participant did not appear in different subsets.

\section{Methods}

\subsection{Overview}
\begin{figure*}[htbp]
  \centering
  \includegraphics[width=0.8\linewidth]{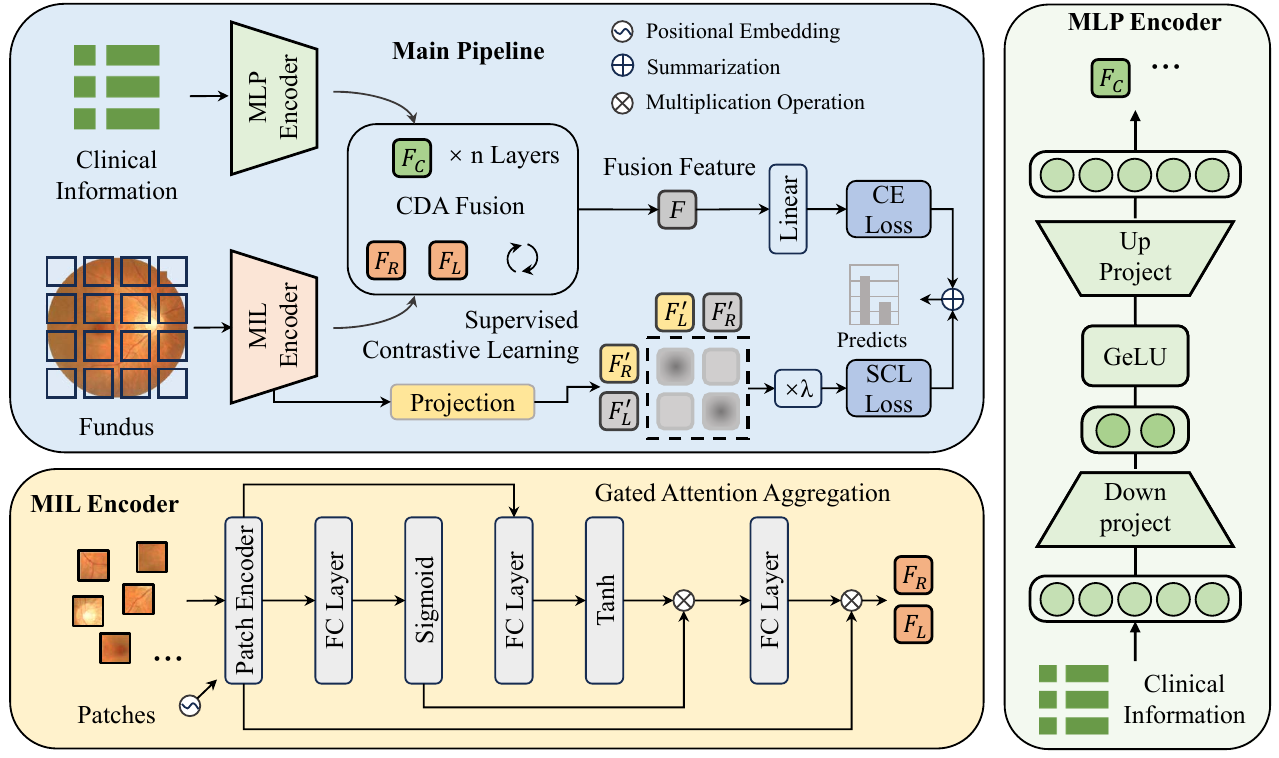} 
  \caption{Overview of the IDNet framework, where the blue section represents the overall workflow encompassing the entire processing chain from data input to output screening results; the orange-yellow section provides a detailed diagram of the MIL Encoder, outlining the feature extraction methodology for Fundus Patches; the green section details the specific architectural design of the MLP Encoder.}
  \label{fig:overview} 
\end{figure*}

IDNet integrates binocular fundus photographs and minimal clinical covariates for IHD screening. As shown in Fig.~\ref{fig:overview}, the framework first extracts eye-level visual representations using sliding-window tiling and MIL, and then fuses binocular and clinical information with the proposed Cross-Modal Distillation Aggregator (CDA). This design addresses the scale mismatch between high-resolution retinal images and low-dimensional tabular inputs while preserving clinically relevant visual details.

\subsection{Sliding Window Tiling and Multi-Instance Learning}
\label{subsec:swt_mil}

Given a high-resolution monocular fundus image $\mathbf{I} \in \mathbb{R}^{C \times H \times W}$, we employ a Sliding Window Tiling (SWT) strategy to decompose the pixel plane into a sequence of overlapping tiles. These tiles are mapped to a feature space via a shared visual backbone and a linear projection layer. Subsequently, a Gated Attention mechanism performs Multi-Instance Learning (MIL) aggregation to yield a unified monocular global vector $\mathbf{v} \in \mathbb{R}^D$. This vector serves as the specific visual input for the subsequent IDNet multi-modal fusion module (yielding $\mathbf{v}_L$ and $\mathbf{v}_R$ for the left and right eyes, respectively).

\subsubsection{Sliding Window Tiling and Feature Extraction}
Let the square window size be $T$ and the overlap ratio be $\rho \in [0, 1)$. The sliding stride $S$ is calculated as:
\begin{equation}
    S = \lfloor T(1 - \rho) \rfloor
\end{equation}
The dimensions of the resulting tile grid, denoted as $n_h$ and $n_w$, are determined by the floor function:
\begin{equation}
    n_h = \left\lfloor \frac{H - T}{S} \right\rfloor + 1, \quad n_w = \left\lfloor \frac{W - T}{S} \right\rfloor + 1
\end{equation}
This results in a total of $N = n_h n_w$ tiles. We extract all tiles and rearrange them into a batch tensor of shape $[B, N, C, T, T]$ (where $B$ represents the batch size), which is then fed into the visual backbone. Each tile is first resized to a fixed resolution (e.g., $224 \times 224$ pixels in this study) to align with the input requirements of the visual encoder $\mathcal{E}$ (utilizing the pre-trained \textbf{RETFound} model). The encoder outputs a feature vector $\mathbf{h}_i \in \mathbb{R}^{D_b}$ for each tile, which is subsequently projected to a unified embedding dimension $D$ via a linear layer with weights $\mathbf{W}_p$:
\begin{equation}
    \mathbf{x}_i = \mathbf{h}_i \mathbf{W}_p
\end{equation}
The sequence of tiles is then flattened in row-major order to form the instance bag matrix $\mathbf{X} = [\mathbf{x}_1, \dots, \mathbf{x}_N]^\top \in \mathbb{R}^{N \times D}$.

To preserve spatial structural information crucial for lesion localization, we incorporate learnable positional embeddings using an additive decomposition strategy. For a tile located at grid coordinate $(u, v)$, the position vector is defined as the sum of a row embedding $\mathbf{P}_{u}^{(h)}$ and a column embedding $\mathbf{P}_{v}^{(w)}$:
\begin{equation}
    \mathbf{p}_{u,v} = \mathbf{P}_{u}^{(h)} + \mathbf{P}_{v}^{(w)}
\end{equation}
The final instance feature is given by $\mathbf{\tilde{x}}_{u,v} = \mathbf{x}_{u,v} + \mathbf{p}_{u,v}$. This decomposition ensures that the parameter complexity and computational cost scale linearly with $n_h + n_w$, leading to stable and efficient training.

\subsubsection{Gated Attention MIL Aggregation}
To aggregate the set of instance features into a single patient-level descriptor while offering interpretability, we employ Gated Attention MIL. For each instance (tile), we compute a gating vector via two parallel Multi-Layer Perceptron (MLP) branches employing $\tanh$ and $\text{sigmoid}$ activations, respectively. These are then projected by a linear layer to produce a scalar score, which is normalized via softmax to obtain the attention weight $\alpha_i$. 
\begin{equation}
\begin{split}
    \alpha_i = \text{softmax}_i \Big( \mathbf{w}_{g}^\top \big[ &\tanh(\mathbf{\tilde{x}}_i \mathbf{W}_v + \mathbf{b}_v) \odot \\
    &\sigma(\mathbf{\tilde{x}}_i \mathbf{W}_u + \mathbf{b}_u) \big] + b_g \Big)
\end{split}
\end{equation}
\begin{equation}
    \mathbf{v} = \sum_{i=1}^{N} \alpha_i \mathbf{\tilde{x}}_i
\end{equation}
Here, $\odot$ denotes element-wise multiplication, and $\sigma$ is the sigmoid function. The gating mechanism allows for the learning of non-linear relationships among instances, realizing a weighted sum where the weights correspond to the relevance of each element. Crucially, the learned attention weights $\alpha_i$ can be directly back-projected to the original image coordinates, serving as a heatmap to provide interpretability by highlighting lesion areas relevant to the diagnosis.

\subsection{Cross-Modal Distillation Aggregator (CDA)}
\label{subsec:cda}

The \textbf{Cross-Modal Distillation Aggregator (CDA)} serves as the core fusion engine of IDNet. Designed as a transferable, plug-and-play module, it enhances multi-modal interaction by addressing a critical challenge in medical data fusion: \textbf{dimensionality mismatch}. High-resolution retinal features are high-dimensional and fine-grained, whereas clinical covariates are sparse and low-dimensional. Naive concatenation or parallel attention often leads to \textbf{feature domination}, where visual signals overwhelm clinical priors. To resolve this, the CDA employs a sequential interaction mechanism, utilizing learnable queries to progressively distill visual evidence under the guidance of clinical information, thereby acting as an effective information bottleneck.

\subsubsection{Preliminaries and Inputs}
For each subject, the input consists of high-resolution binocular fundus features and clinical tabular data (age, gender, smoking, alcohol history). Following the upstream SWT-MIL processing, we obtain monocular feature vectors for the left and right eyes, denoted as $\mathbf{v}_L, \mathbf{v}_R \in \mathbb{R}^{B \times D}$. Similarly, the clinical covariates are encoded via a Multi-Layer Perceptron (MLP) into a feature vector $\mathbf{\tilde{c}} \in \mathbb{R}^{B \times D}$.

To facilitate sequence-based interaction, we treat these features as single-token sequences. We define the modality-specific input tensors as:
\begin{equation}
\begin{aligned}
    \mathbf{V}_L &= [\mathbf{v}_L] \in \mathbb{R}^{B \times 1 \times D}, \\
    \mathbf{V}_R &= [\mathbf{v}_R] \in \mathbb{R}^{B \times 1 \times D}, \\
    \mathbf{C} &= [\mathbf{\tilde{c}}] \in \mathbb{R}^{B \times 1 \times D}
\end{aligned}
\end{equation}
Crucially, we initialize a set of learnable parameters known as \textit{Fusion Queries}, denoted by a shared latent tensor $\mathbf{S}^{(0)} \in \mathbb{R}^{N_q \times D}$ (where $N_q$ is the number of query tokens). This latent tensor is broadcast across the batch dimension $B$ during training, serving as the anchor for aggregating information.

\subsubsection{Architecture and Iterative Fusion}
\begin{figure}[htbp]
  \centering
  \includegraphics[width=1\linewidth]{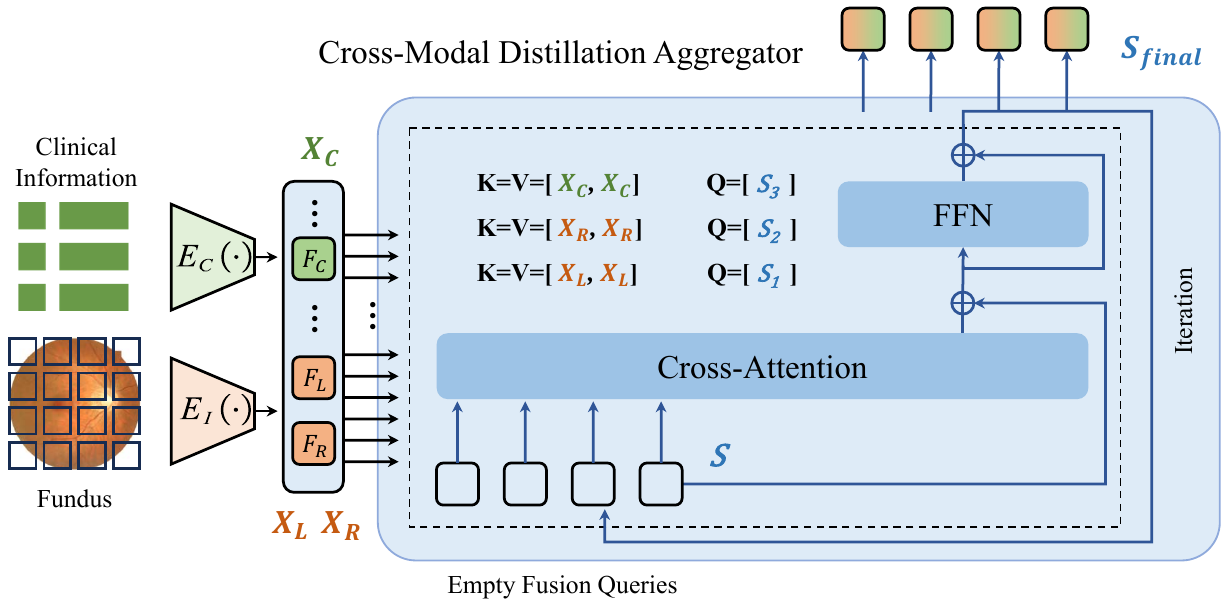} 
  \caption{Detailed schematic of the Cross-Modal Distillation Aggregator showing the iterative query updates.}
  \label{fig:CDA} 
\end{figure}

The CDA architecture (Fig.~\ref{fig:CDA}) is composed of stacked interaction layers. Within each layer, the shared latent $\mathbf{S}$ acts as a query probe, iteratively interacting with the Left Eye, Right Eye, and Clinical modalities via a Cross-Attention mechanism. This design ensures a structured and hierarchical distillation of information.

Each interaction step employs a standard Transformer block structure with Pre-Normalization and Residual Connections. Taking the \textit{Left Eye Interaction} as an example, let the current latent state be $\mathbf{S}$. We define the query ($\mathbf{Q}$), key ($\mathbf{K}$), and value ($\mathbf{V}$) matrices as:
\begin{equation}
    \mathbf{Q} = \text{LN}(\mathbf{S}), \quad \mathbf{K} = \mathbf{V}_L, \quad \mathbf{V} = \mathbf{V}_L
\end{equation}
where $\text{LN}(\cdot)$ denotes Layer Normalization. The interaction is computed using Multi-Head Attention (MHA):
\begin{equation}
    \text{MHA}(\mathbf{Q}, \mathbf{K}, \mathbf{V}) = \text{Concat}(\text{head}_1, \dots, \text{head}_H)\mathbf{W}_O
\end{equation}
\begin{equation}
    \text{head}_h = \text{softmax}\left( \frac{(\mathbf{Q}\mathbf{W}_Q^{(h)})(\mathbf{K}\mathbf{W}_K^{(h)})^\top}{\sqrt{d_h}} \right) (\mathbf{V}\mathbf{W}_V^{(h)})
\end{equation}
Here, $H$ is the number of attention heads and $d_h = D/H$ is the dimension per head. The latent state is updated via residual connections followed by a Feed-Forward Network (FFN):
\begin{equation}
\begin{aligned}
    \mathbf{\tilde{S}} = \mathbf{S} + \text{MHA}(\text{LN}(\mathbf{S}), \mathbf{V}_L, \mathbf{V}_L),\\ \quad \mathbf{S}' = \mathbf{\tilde{S}} + \text{FFN}(\text{LN}(\mathbf{\tilde{S}}))
\end{aligned}
\end{equation}
The FFN consists of two linear layers with a GELU activation: $\text{Linear}(D, 4D) \to \text{GELU} \to \text{Linear}(4D, D)$.

Following the left eye, the updated latent $\mathbf{S}'$ sequentially interacts with the \textit{Right Eye} features ($\mathbf{V}_R$) to produce $\mathbf{S}''$:
\begin{equation}
\begin{aligned}
    \mathbf{\tilde{S}}' = \mathbf{S}' + \text{MHA}(\text{LN}(\mathbf{S}'), \mathbf{V}_R, \mathbf{V}_R),\\ \quad \mathbf{S}'' = \mathbf{\tilde{S}}' + \text{FFN}(\text{LN}(\mathbf{\tilde{S}}'))
    \end{aligned}
\end{equation}
Finally, the latent $\mathbf{S}''$ interacts with the \textit{Clinical} features ($\mathbf{C}$) to complete one full CDA layer cycle:
\begin{equation}
\begin{aligned}
    \mathbf{\tilde{S}}'' = \mathbf{S}'' + \text{MHA}(\text{LN}(\mathbf{S}''), \mathbf{C}, \mathbf{C}),\\ \quad \mathbf{S}^{(+)} = \mathbf{\tilde{S}}'' + \text{FFN}(\text{LN}(\mathbf{\tilde{S}}''))
\end{aligned}
\end{equation}

This sequential process can be formally expressed as a composition of modality-specific update functions $\Phi$. For a CDA module with $T$ stacked layers, the output at layer $t$ is:
\begin{equation}
\begin{aligned}
    \mathbf{S}^{(t)} = \Phi_{\text{clin}} \circ \Phi_{\text{right}} \circ \Phi_{\text{left}} (\mathbf{S}^{(t-1)}),\\ \quad t = 1, \dots, T
\end{aligned}
\end{equation}
where each $\Phi$ encapsulates the "Pre-Norm MHA + Residual + Pre-Norm FFN + Residual" block.

The final fused representation $\mathbf{z}$ is obtained by pooling the output of the last layer $\mathbf{S}^{(T)}$:
\begin{equation}
    \mathbf{z} = \text{Pool}(\mathbf{S}^{(T)}) \in \mathbb{R}^D
\end{equation}
By decoupling the fusion mechanism from the specific visual encoder, the CDA functions as a versatile module that can be easily transferred to other architectures, providing stable training dynamics and effective regularization while preserving interpretability.

\subsection{Learning Objective}
\label{subsec:loss}

We propose a joint optimization framework combining discriminative classification with robust representation learning. The total loss $\mathcal{L}$ is formulated as:
\begin{equation}
    \mathcal{L} = \mathcal{L}_{\mathrm{cls}} + \lambda_{\mathrm{con}} \mathcal{L}_{\mathrm{SupCon}}
\end{equation}
where $\lambda_{\mathrm{con}}$ is a hyperparameter balancing the contrastive objective.

\subsubsection{Focal Cross-Entropy Loss}
To mitigate the impact of class imbalance and easy negatives, we employ \textbf{Focal Loss}~\cite{lin2017focal}. By introducing a focusing parameter $\gamma$ and a weighting factor $\alpha$, the loss is defined as:
\begin{equation}
    \mathcal{L}_{\mathrm{cls}} = -\alpha_y (1 - \hat{p}_y)^\gamma \log \hat{p}_y
\end{equation}
where $\hat{p}_y$ is the predicted probability for the target class. The term $(1 - \hat{p}_y)^\gamma$ effectively down-weights well-classified examples, forcing the model to focus optimization on hard-to-classify samples.

\subsubsection{Supervised Contrastive Loss (SCL)}
We explicitly model binocular consistency by treating the left and right eye images of the same subject as positive pairs. The Supervised Contrastive Loss~\cite{khosla2020supervised} is applied to the normalized feature vectors $\mathbf{\bar{z}}$:
\begin{equation}
    \mathcal{L}_{\mathrm{SupCon}} = \sum_{i \in I} \frac{-1}{|\mathcal{P}(i)|} \sum_{p \in \mathcal{P}(i)} \log \frac{\exp(\langle \mathbf{\bar{z}}_i, \mathbf{\bar{z}}_p \rangle / \tau)}{\sum_{a \in \mathcal{A}(i)} \exp(\langle \mathbf{\bar{z}}_i, \mathbf{\bar{z}}_a \rangle / \tau)}
\end{equation}
Here, $\mathcal{P}(i)$ denotes the set of positive samples for anchor $i$, and $\tau$ is the temperature parameter. This objective enforces the alignment of patient-specific binocular features in the embedding space while maximizing the distance between distinct subjects, thereby enhancing representation robustness.

\begin{table*}[htbp]
  \centering
  \caption{Performance comparison of IDNet with Baselines and SOTA models on the UKB dataset.}
  \label{tab:sota_comparison}
  \small
  \begin{tabular}{lllc}
    \toprule
    \textbf{Category} & \textbf{Models} & \textbf{Key Components} & \textbf{ROC-AUC} \\
    \midrule
    \multirow{2}{*}{Baselines} 
      & Clinical-Only & MLP Encoder & 0.7601 \\
      & Image-Only & RETFound + SWT + MIL + SCL & 0.7991 \\
    \midrule
    Naive Fusion & BaseModel (Concat) & RETFound (origin) + MLP + Concat & 0.7864 \\
    \midrule
    \multirow{5}{*}{SOTA Strategies} 
      & MIL-ViT\cite{bi2023mil} + CDA & ViT + MIL Head + CDA & 0.7989 \\
      & MSTNet\cite{wei2025mstnet} + CDA & Multi-scale Spatial-aware Transformer + CDA & 0.8009 \\
      & MCAT\cite{chen2021multimodal} & RETFound + SWT + MIL + SCL + MCAT & 0.8012 \\
      & MOTCAT\cite{xu2023multimodal} & RETFound + SWT + MIL + SCL + MOTCAT & 0.8029 \\
      & HealNet\cite{hemker2024healnet} & RETFound + SWT + MIL + SCL + HealNet & 0.8066 \\
    \midrule
    \textbf{Ours} & \textbf{IDNet (CDA)} & \textbf{RETFound + SWT + MIL + SCL + CDA} & \textbf{0.8168} \\
    \bottomrule
  \end{tabular}
\end{table*}

\section{Experiment}

\subsection{Implementation Details}
We implemented the proposed framework using PyTorch on an Ubuntu 22.04 operating system, utilizing an Nvidia A100 GPU for both training and testing. The IHD screening model was trained with a learning rate of $5 \times 10^{-4}$, employing an early stopping strategy that typically reached optimal performance around epoch 24. We set the weight decay to $1 \times 10^{-3}$ and the batch size to 16. For the joint loss function, the weighting hyperparameter $\lambda_{\mathrm{con}}$ for the Supervised Contrastive Loss was set to 0.2. To ensure fair comparison, these implementation parameters were kept consistent across all control experiments and the subsequent ablation studies.

\subsection{Comparative Analysis}

To rigorously validate the effectiveness of IDNet, we conducted a comprehensive comparison against mainstream visual encoders and state-of-the-art (SOTA) fusion strategies on the UKB dataset. We evaluated performance using Sensitivity, ROC-AUC, and F1-score.

\begin{table}[htbp]
  \centering
  \caption{Classification performance of different methods.}
  \label{tab:main_comparison}
  \setlength{\tabcolsep}{1pt}
  \begin{tabular}{llccc}
    \toprule
    \textbf{Models} & \textbf{Fusion Methods} & \textbf{Sensitivity} & \textbf{ROC-AUC} & \textbf{F1-score} \\
    \midrule
    ResNet-101\cite{he2016deep} & Concatenation & 0.5528 & 0.7589 & 0.3965 \\
    ViT\cite{dosovitskiy2020image} & Concatenation & 0.5862 & 0.7541 & 0.4012 \\
    VisionRWKV\cite{duan2024vision} & Concatenation & 0.5894 & 0.7659 & 0.4325 \\
    VisionMamba\cite{liu2025vision} & Concatenation & 0.5862 & 0.7751 & 0.4423 \\
    Swin Transformer\cite{liu2021swin} & Concatenation & 0.6101 & 0.7759 & 0.4535 \\
    ConvNeXt\cite{woo2023convnext} & Concatenation & 0.5922 & 0.7676 & 0.4421 \\
    RETFound\cite{zhou2023foundation} & Concatenation & 0.6012 & 0.7864 & 0.4533 \\
    RETFound\cite{zhou2023foundation} & Late Fusion & 0.6041 & 0.7789 & 0.4456 \\
    \midrule
    \textbf{IDNet (Ours)} & \textbf{CDA} & \textbf{0.6311} & \textbf{0.8168} & \textbf{0.4682} \\
    \bottomrule
  \end{tabular}
\end{table}

Tables~\ref{tab:sota_comparison} and~\ref{tab:main_comparison} show that simple multimodal fusion does not reliably improve performance. In particular, direct concatenation of retinal and clinical features yields lower ROC-AUC than the image-only baseline, suggesting that naive fusion is insufficient for this task. In contrast, IDNet achieves the best overall performance (ROC-AUC 0.8168), indicating that the combination of high-resolution visual modeling and CDA-based fusion is more effective than standard concatenation and late-fusion baselines.

\subsection{Ablation Studies}

To systematically quantify the contributions of the proposed components, we performed ablation studies on the UKB test set (Table~\ref{tab:ablation}). The \textbf{BaseModel}, which naively concatenates RETFound features with clinical data, achieved an ROC-AUC of 0.7864, constrained by the loss of detail from image resizing.

\begin{table}[htbp]
  \centering
  \caption{Ablation study on the importance of individual components.}
  \label{tab:ablation}
  \setlength{\tabcolsep}{3pt}
  \begin{tabular}{lcccccc}
    \toprule
    \multirow{2}{*}{\textbf{Models}} & \multicolumn{4}{c}{\textbf{Components}} & \multicolumn{2}{c}{\textbf{Metrics}} \\
    \cmidrule(lr){2-5} \cmidrule(lr){6-7}
     & \textbf{MIL} & \textbf{SWT} & \textbf{SCL} & \textbf{CDA Fusion} & \textbf{ROC-AUC} & \textbf{F1-score} \\
    \midrule
    BaseModel & $\times$ & $\times$ & $\times$ & $\times$ & 0.7864 & 0.4533 \\
    Ours-1    & \checkmark & $\times$ & $\times$ & $\times$ & 0.8006 & 0.4596 \\
    Ours-2    & \checkmark & \checkmark & $\times$ & $\times$ & 0.8045 & 0.4613 \\
    Ours-3    & \checkmark & \checkmark & \checkmark & $\times$ & 0.8085 & 0.4611 \\
    \textbf{IDNet} & \checkmark & \checkmark & \checkmark & \checkmark & \textbf{0.8168} & \textbf{0.4682} \\
    \bottomrule
  \end{tabular}
\end{table}

The progressive integration of our modules yielded consistent gains. Incorporating Attention-based MIL (\textbf{Ours-1}) improved the AUC to 0.8006 by adaptively aggregating instance features. Adding the Sliding Window Tiling strategy (\textbf{Ours-2}) further boosted performance to 0.8045, proving that preserving high-resolution details via tiling is crucial for capturing localized lesions. Furthermore, introducing Supervised Contrastive Loss (\textbf{Ours-3}) raised the AUC to 0.8085. By enforcing feature alignment between the left and right eyes of the same patient, SCL enhances the robustness and generalizability of the learned representations. Finally, replacing naive concatenation with our IDNet Fusion module (\textbf{CDA}) delivered the most significant leap, achieving a final AUC of 0.8168. Unlike standard Late Fusion, the CDA effectively manages multi-modal interactions, validating its superiority in distilling heterogeneous information.

\subsection{Generalization of CDA (Plug-and-Play)}

To verify the versatility and ``plug-and-play'' capability of the proposed Cross-Modal Distillation Aggregator (CDA), we integrated it into a diverse set of visual backbones, ranging from standard architectures (ViT, Swin Transformer) to emerging state-space models (VisionMamba, VisionRWKV) and specialized CFP feature extraction screening networks (MIL-ViT, MSTNet).

\begin{table}[htbp]
  \centering
  \caption{Performance improvement with CDA across different visual encoders.}
  \label{tab:plug_and_play}
  \setlength{\tabcolsep}{12pt}
  \begin{tabular}{lcc}
    \toprule
    \multirow{2}{*}{\textbf{Vision Encoder}} & \multicolumn{2}{c}{\textbf{Fusion Strategy (ROC-AUC)}} \\
    \cmidrule(lr){2-3}
     & \textbf{Without CDA} & \textbf{With CDA} \\
    \midrule
    ViT\cite{dosovitskiy2020image} & 0.7541 & 0.7752 \textbf{(+2.11\%)} \\
    VisionRWKV\cite{duan2024vision} & 0.7659 & 0.7704 \textbf{(+0.45\%)} \\
    VisionMamba\cite{liu2025vision} & 0.7751 & 0.7813 \textbf{(+0.62\%)} \\
    Swin Transformer\cite{liu2021swin} & 0.7759 & 0.7864 \textbf{(+1.05\%)} \\
    MIL-ViT\cite{bi2023mil} & 0.7799 & 0.7989 \textbf{(+1.90\%)} \\
    MSTNet\cite{wei2025mstnet} & 0.7914 & 0.8009 \textbf{(+0.95\%)} \\
    \bottomrule
  \end{tabular}
\end{table}

As detailed in Table~\ref{tab:plug_and_play}, the integration of CDA consistently improved classification performance across all tested architectures. For instance, the standard Vision Transformer saw a significant boost of 2.11\%, mitigating the limitations of simple feature aggregation. Even for advanced domain-specific models like MSTNet, which already possess strong feature extraction capabilities (0.7914 AUC), the CDA module provided a further gain of 0.95\%, pushing the performance to 0.8009. These consistent improvements demonstrate that the CDA is not merely a model-specific component, but a robust, generalized fusion engine. It effectively distills complementary clinical information to refine visual features, regardless of the underlying encoder architecture, thereby validating its value as a universal plug-and-play module for multi-modal medical image analysis.

\subsection{Interpretability Assessment}

\begin{figure}[htbp]
  \centering
  \includegraphics[width=1\linewidth]{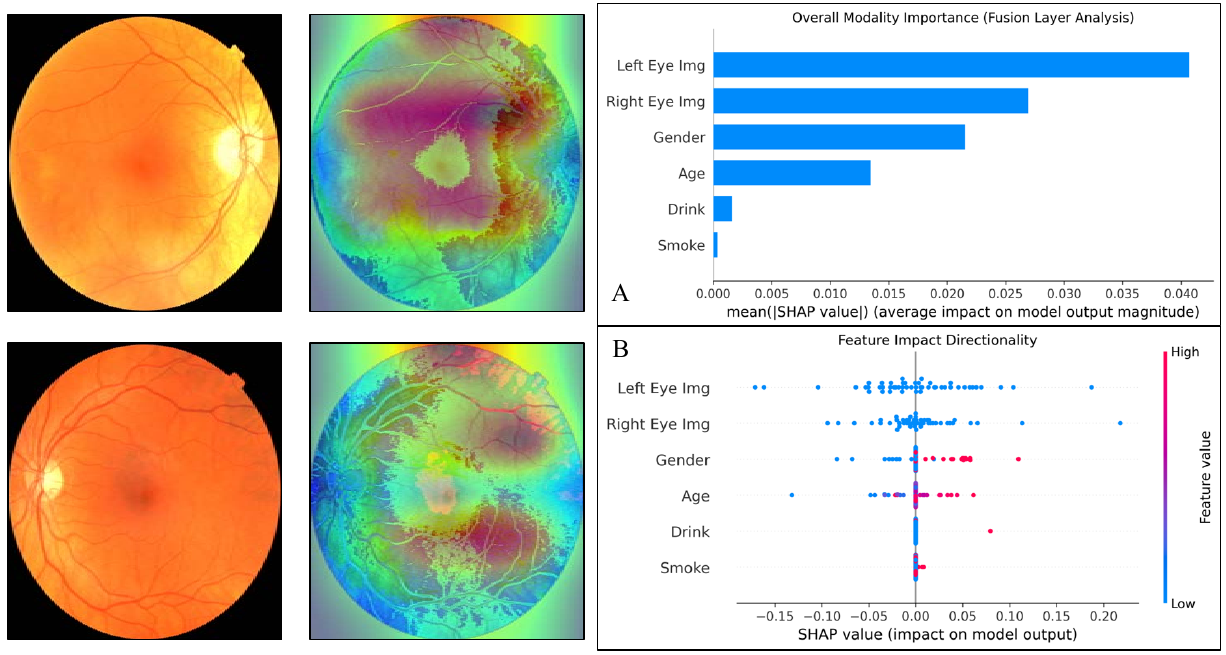} 
  \caption{Interpretability Analysis. Saliency maps (Left) highlight the model's focus on retinal vascular morphology. SHAP analysis (Right) confirms: (A) The dominance of retinal features over clinical data; (B) Age and gender correlate correctly with risk, while lifestyle impacts are implicitly encoded in visual features.}
  \label{fig:explainable} 
\end{figure}

As shown in Fig.~\ref{fig:explainable}, the model focuses mainly on retinal vascular regions, which is consistent with the vascular pathology of IHD. The SHAP analysis also suggests that retinal features contribute more strongly than the four clinical covariates, while age and sex remain informative auxiliary factors. These observations support the clinical plausibility of the learned representation and are consistent with the quantitative results.

\subsection{External Validation in Real-World Screening Scenarios}
We further evaluated IDNet on an independent external cohort of 3,054 subjects (850 IHD-positive and 2,204 controls), using binocular fundus photographs and the same four clinical covariates. The external images were acquired at a higher resolution ($6960\times4640$) than those in UK Biobank ($2048\times1536$). On this cohort, IDNet achieved a ROC-AUC of 0.8965, with sensitivity, specificity, accuracy, and F1-score of 0.9247, 0.7518, 0.7999, and 0.7201, respectively, at an operating threshold of 0.41. Notably, the robust performance on this external cohort can be partially attributed to the superior acquisition quality of the images. The higher-fidelity fundus photographs likely provided clearer depictions of critical retinal microvascular biomarkers associated with IHD, which IDNet effectively captured. The result suggests that the model transfers well to an independent real-world screening setting. The external cohort was collected independently and was not used for model development, hyperparameter tuning, or model selection. The operating threshold was selected on the validation set and then fixed for external evaluation.

\section{Conclusion}
\label{sec:conclusion}

In this work, we presented \textbf{IDNet}, a novel and unified multi-modal framework designed for accessible Ischemic Heart Disease (IHD) screening. By synergizing high-resolution fundus imaging with minimal clinical covariates via the proposed \textbf{Cross-Modal Distillation Aggregator (CDA)}, our method effectively overcomes the limitations of existing single-modal baselines and naive fusion approaches. The CDA, functioning as a versatile plug-and-play module, demonstrates robust generalization across various visual encoders. Crucially, extensive external validation on an independent, real-world screening cohort with ultra-high-resolution images yielded an outstanding ROC-AUC of 0.8965, further confirming the framework's clinical applicability and system robustness. Furthermore, we constructed a large-scale, high-quality benchmark dataset, bridging a critical gap in resources for IHD research and facilitating the development of data-driven screening algorithms.

\section*{Acknowledgment}
This research has been conducted using the UK Biobank Resource under Application Number 360934. The authors extend their sincere gratitude to Prof. Jihong Wu (Department of Ophthalmology, Eye and ENT Hospital, Fudan University; Shanghai Key Laboratory of Visual Impairment and Restoration; and Key Laboratory of Myopia and Related Eye Diseases, NHC, Shanghai, China) for securing the UK Biobank data access and supporting this work as a preliminary outcome of his overarching project. Furthermore, the authors would like to thank the participants and staff of the UK Biobank for their dedication and contribution to the resource.

\bibliographystyle{IEEEbib}
\bibliography{SMC2026references}

\end{document}